\documentclass[twoside,11pt]{article}
\usepackage{jmlr2e}
\usepackage{algorithm}
\usepackage[algo2e,ruled,vlined,linesnumbered]{algorithm2e}

\usepackage{amsfonts}

\newcommand{\set}[1]{\left\{ #1\right\}}

\newcommand{\sodass}{\,:\,}
\newcommand{\setGilt}[2]{\left\{ #1\sodass #2\right\}}

\newcommand{\realrange}[2]{\left[#1, #2\right]}

\newcommand{\unitrange}[2]{\realrange{0}{1}}

\newcommand{\llabel}[1]{\label{\labelprefix:#1}}
\newcommand{\labelprefix}{} 

\newcommand{\discussionsize}{\small}

\newcommand{\frage}[1]{}

\newenvironment{code}{\noindent
\begin{tabbing}%
\hspace{2em}\=\hspace{2em}\=\hspace{2em}\=\hspace{2em}\=\hspace{2em}\=%
\hspace{2em}\=\hspace{2em}\=\hspace{2em}\=\hspace{2em}\=\hspace{2em}\=%
\kill}{\end{tabbing}}

\newcommand{\labelcommand}{}
\newcommand{\captiontext}{}
\newsavebox{\codeparam}
\newcounter{lineNumber}
\newenvironment{disscodepos}[3]{%
\renewcommand{\labelcommand}{#2}%
\renewcommand{\captiontext}{#3}%
\sbox{\codeparam}{\parbox{\textwidth}{#3}}%
\begin{figure}[#1]\begin{center}\begin{code}\setcounter{lineNumber}{1}}{%
\end{code}\end{center}\caption{\llabel{\labelcommand}\captiontext}\end{figure}}

{\end{disscodepos}}

\newcommand{\Is}       {:=}

\newdimen\endofsize\endofsize=0.5em
\def\endofbeweis{~\quad\hglue\hsize minus\hsize
                 \hbox{\vrule height \endofsize width
\endofsize}\par}

\usepackage{multirow}
\usepackage{numprint}

\usepackage{amsmath}
\usepackage{hyperref}
\usepackage{url}

\usepackage[usenames,dvipsnames]{xcolor}
\usepackage[normalem]{ulem}
\usepackage[]{graphicx}\usepackage[]{color}
\def\MdR{\ensuremath{\mathbb{R}}}

\newcommand{\etal}{et~al.~}
\newcommand{\eg}{e.g.\ }

\usepackage{capt-of}%
\usepackage{caption}

\usepackage{etoolbox}
\usepackage[left,pagewise] {lineno}
\definecolor {infocolor} {rgb} {0.6,0.6,0.6}
\definecolor{bole}{rgb}{0.47, 0.27, 0.23}
\patchcmd{\thebibliography}{\list}{\fontsize{0.98em}{0.9\baselineskip}\selectfont\list}{}{} 

\begin{document}
\title{\Large Faster Support Vector Machines\thanks{This work was partially supported by DFG SA 933/11-1, SCHU 2567/1-2 and the research leading to these results has received funding from the European Research Council under the European Union's Seventh Framework Programme (FP/2007-2013) / ERC Grant Agreement no. 340506.}}
\author{Sebastian Schlag\thanks{Institute for Theoretical Informatics, Karlsruhe Institute of Technology, Karlsruhe, Germany.} \\
   \and
   Matthias Schmitt\thanks{Karlsruhe Institute of Technology, Karlsruhe, Germany.} \\
   \and
   Christian Schulz\thanks{University of Vienna, Faculty of Computer Science, W\"ahringer Str. 29, 1080 Vienna, Austria.}}

\date{}

\maketitle

\begin{abstract} \small\baselineskip=9pt
  The time complexity of support vector machines (SVMs) prohibits training on huge data sets with millions of data points.
  Recently, multilevel approaches to train SVMs have been developed to allow for time-efficient training on huge data sets.
  While regular SVMs perform the entire training in one -- time consuming -- optimization step, multilevel SVMs first
  build a hierarchy of problems decreasing in size that resemble the original problem and then train an SVM model for each hierarchy level, benefiting from the solved models of previous~levels.
  We present a faster multilevel support vector machine that uses a label propagation algorithm to construct the problem hierarchy. 
Extensive experiments indicate that our approach is up to orders of magnitude faster than the previous fastest algorithm while having comparable~classification~quality.
For example, already one of our sequential solvers is on average a factor 15 faster than the parallel ThunderSVM algorithm, while having similar classification quality.
\end{abstract}

\section{Introduction}
Machine learning is an important subfield of computer science
that builds and studies algorithms which are able to learn from and to understand the
vast amounts of data that are available today in order to make predictions.
A concrete machine learning task is the classification problem.
In a classification problem, we are given unlabeled data points, \eg information about
the financial situation of a person, and want to put them into the right class
out of a finite number of classes, e.g credit-worthy or not credit-worthy.
Such tasks were historically done by experts in the specific field and required
lots of time and man power.
We can easily see why a bank would like to automate the process of checking for
credit-worthiness because a machine could do this job faster, more cost-efficiently, and hopefully less error prone.
With recent advances in machine learning, we now have algorithms that are able to
do the work previously thought of being an exclusive competence of humans.
The training is done by presenting the algorithm with labeled example data
points from which it has to learn the underlying structure such that it
is able to correctly classify unlabeled data afterwards.

Large margin classifiers are one approach to tackle this problem.
These classifiers separate the classes of the classification problem in
such a way that they are able to give the distance to a decision boundary.
Support Vector Machines (SVMs)~\citep{cortes1995support} are the most well-known large margin classifiers.
They solve a convex optimization problem and use a maximum-margin hyperplane to
separate classes.
SVMs are known to show good performance when trained to solve a classification
problem, but in order to achieve high quality predictions \emph{model selection} is also
required.
Model selection is the process of finding the right parameters for a specific problem.
This is the work-intensive part of machine learning
with SVMs.
Since the time complexity for solving the optimization problem underlying SVMs
is between $O(n^2)$ and $O(n^3)$~\citep{DBLP:conf/nips/GrafCBDV04}, and every set of parameters requires the
training of a new SVM model, SVM training becomes a problem
on data sets that have hundreds of thousands or even millions of training points.

Model selection itself is highly parallelizable as different model parameters can be evaluated independently in parallel, 
but for large data sets the time complexity to solve the underlying optimization problem for a single set of parameters is still infeasible.
Other model selection approaches for SVM make some training results reusable~\citep{DBLP:journals/ml/ChapelleVBM02}
so that the training time complexity shrinks but parallelization is more
difficult. Even highly optimized SVM algorithms can not cope with data
sets of hundreds of~thousands~of~data~points.

In practice, huge data sets can be imbalanced, i.e., classes have unequal size.
An example are medical data sets where labels tell whether a person is ill or 
healthy.
A data set like this is often  
imbalanced due to the fact that illnesses occur less often when the general
population is considered.
Different machine learning algorithms apply varying techniques to train SVMs on such data sets.

One possibility to cope with large data sets is random sampling~\citep{schohn2000less}. Here, a random subset of the input data is selected and used for
training. However, problems can arise when infrequently occurring important data are underrepresented in the sample.
Because random sampling only reflects the distribution of training data one may 
miss significant regions of the testing data~\citep{Yu03}.
Hence, to train a SVM on large problems a more sophisticated approach is needed.

A new promising research path to tackle the scalability problem was recently introduced to SVMs: \emph{the multilevel paradigm}.
Already widely used in, e.g. graph partitioning~\citep{SPPGPOverviewPaper}, a multilevel framework first
builds a hierarchy of the problem.
Each level of the hierarchy is a problem instance that decreases in size but reflects the structure of the original problem.
Then, a regular SVM is trained on the coarsest problem.
This is more
feasible than training on the original input because the problem is much smaller.
The training is then projected upwards
in the hierarchy, i.e., the model is refined to better fit the original problem while
only gradually increasing the size of the data the SVM is trained on.
Overall, experiments indicate that the multilevel paradigm reduces computation time while being comparable and often better
than non-hierarchical approaches in terms of prediction quality on large data~sets. 

\paragraph{Contribution and Outline.}
\label{sec:contribution}
We present a faster SVM that uses the multilevel paradigm. 
In contrast to approaches such as LibSVM~\citep{libsvm} or ThunderSVM~\citep{wenthundersvm18} that perform expensive model selection on the input instance, using the multilevel paradigm enables us to perform model selection very quickly on a much smaller instance and afterwards to perform fast training. 
Multilevel approaches in this context have been introduced by \cite{razzaghi15}. In contrast to their approach, we use a clustering-based contraction scheme that is able to shrink the input size very quickly and hence to build a multilevel hierarchy much quicker.
More precisely, we apply a near-linear~time label propagation clustering algorithm to compute a clustering and then contract the clustering.
In this extended version of the paper we add two important algorithms. First, since the bottleneck in our multilevel algorithm is the training phase, we now include a parallel version of our algorithm that uses a parallel training algorithm. Secondly, we introduce an algorithm that computes clusters having a low-diameter to compute the hierarchy in the multilevel algorithm. Using these clusterings yield training problems during uncoarsening  yield overall improved quality.
Extensive experiments indicate that our approach is up to orders of magnitude faster than the previous fastest algorithm while having comparable~classification~quality.
For example, already one of our sequential solvers is on average a factor 15 faster than the parallel ThunderSVM algorithm, while having similar classification quality.
\section{Preliminaries}

\subsection{Basic Concepts}Let  $G=(V=\{0,\ldots, n-1\},E,\omega)$ be an \emph{undirected graph} 
with edge weights $\omega: E \to \MdR_{>0}$, $n = |V|$, and $m = |E|$.
We extend $\omega$ to sets, i.e.,
 $\omega(E')\Is \sum_{e\in E'}\omega(e)$.
$N(v)\Is \setGilt{u}{\set{v,u}\in E}$ denotes the \emph{neighbors} of $v$.
Given a set of points $P=\{p_1,\ldots,p_m\}$, with $p_i \in \MdR^d$, the \emph{$k$-nearest neighbor graph} has a vertex for every point and connects two vertices $p, q$ by an edge if the distance between them is among the $k$-th smallest distances from $p$ to other points in~$P$.
The \emph{diameter} of a graph is the length of the longest of the shortest path between any pair of vertices.
An \emph{independent set} $I\subset V$ in a graph $G$ is a set of vertices such that no two
vertices are adjacent.
A \emph{clustering} is a partition of the nodes, i.e., \emph{blocks} of nodes $V_1$,\ldots,$V_k \subset V$ 
such that $V_1\cup\cdots\cup V_k=V$ and $V_i\cap V_j=\emptyset$.
 However, $k$ is usually not given in advance. 
A \emph{size-constrained clustering} constrains the size of the blocks by a given upper bound $U$ such that $|V_i| \leq U$. 
For example, when using $U=1$, the only feasible size-con\-strained clustering in an unweighted graph is the singleton clustering, where each node forms a block~on~its~own.  

\subsection{Classification}
A common supervised learning task in machine learning is the
classification problem, i.e., to associate unlabeled data points with a specific
class out of a finite number of classes.
We do so by training an algorithm on a set of training examples \(x_1,\ldots,x_n\) with associated labels $y_{1},\ldots,y_{n}$.
Once the algorithm is trained on the training set, we predict the
class/label $y_{n+1}$ for a new data point $x_{{n+1}}$.
A special case of classification is binary classification, where there are only
two classes. 

\subsection{Support Vector Machines}
\label{sec:svm}
Support vector machines (SVMs) are supervised learning models that can be used
for classification. SVMs  are one of the most
well-known machine learning algorithms~\citep{emlsvm}.
They are large margin classifiers that find a hyperplane to decide
the class for a new data point.

Given a set $\mathcal{I}$ of $n$ data points $x_i$ with corresponding
labels $y_i \in \{+1,-1\}$, the minority class $\mathbf{C}^+$
consists of all data points with positive label ($|\mathbf{C}^+|=n^+$).
All other data points are in the majority class $\mathbf{C}^-$
($|\mathbf{C}^-|=n^-$, $n = n^+ + n^-$).
Throughout this paper we assume w.l.o.g. $n^+ \le n^-$.
Every training data point~$x_i$ is interpreted as a $d$ dimensional vector in
$\mathbb{R}^d$. The SVM finds a $d-1$ dimensional hyperplane
separating the two classes.
The \emph{best separating hyperplane} is the one furthest away from both classes, i.e., the
one with the largest margin between the two classes, hence the name large margin
classifier. 

However, often the data is not linearly separable in the Euclidean space.
Hence, in practice, one uses a mapping of the data points to a higher dimensional space $\phi: \MdR^d \to \MdR^p$ $(d \leq p)$ in order to make two classes separable by a hyperplane.
This is also known as the kernel trick, which was originally proposed by \cite{Aizerman:1964} and applied to SVMs by \cite{Boser92}.
In this paper, we use the Gaussian kernel (radial basis function, RBF), i.e., $k(x_i,x_j) = \exp(-\gamma \lVert x_i - x_j \rVert^2) = \phi(x_i)^{T}\phi(x_j)$.
This function is known to be reliable when no additional assumptions about the
data are known~\citep{Scholkopf_LKS}. The standard SVM formulation is then given by the following
constrained optimization problem:
\begin{align*}
\text{minimize} \qquad & \frac{1}{2} \lVert w \rVert^2 + C \sum^n_{i=1}\xi_i \\
\text{subject to} \qquad & y_i (w \cdot \phi( x_i ) - b) \geq 1 - \xi_i , \quad \xi_i \geq 0 .
\end{align*}

The hyperplane $f(x) = w \cdot \phi(x) + b$ with maximum margin is calculated by
solving for the system and obtaining optimal parameters $w$ and $b$.
The slack variables \(\xi_i = \max(0, 1-y_i(w \cdot \phi(x_i) - b))\) are used to extend SVMs to
cases where data is not linearly separable, meaning there is no hyperplane
separating $\mathbf{C}^+$ and $\mathbf{C}^-$.
This allows for misclassification but ensures that every $x_i$ lies
on the correct side of the margin. Since it is part of the term that is to
minimize, it also penalizes misclassification.
This method is known as soft margin extension~\citep{cortes1995support} to the SVM.
The parameter $C > 0$ controls the magnitude of~the~penalization.

Once the vectors $w$ and $b$ have been found, new data points are classified by the sign of the hyperplane equation
\(h(x) = \text{sign}(w \cdot \phi(x) + b)\).
In simpler terms the hyperplane splits the input space into two and puts data
points above the plane in $\mathbf{C}^+$ and below in $\mathbf{C}^-$.
\begin{figure}[t]
  \centering
  \includegraphics[width=.45\linewidth]{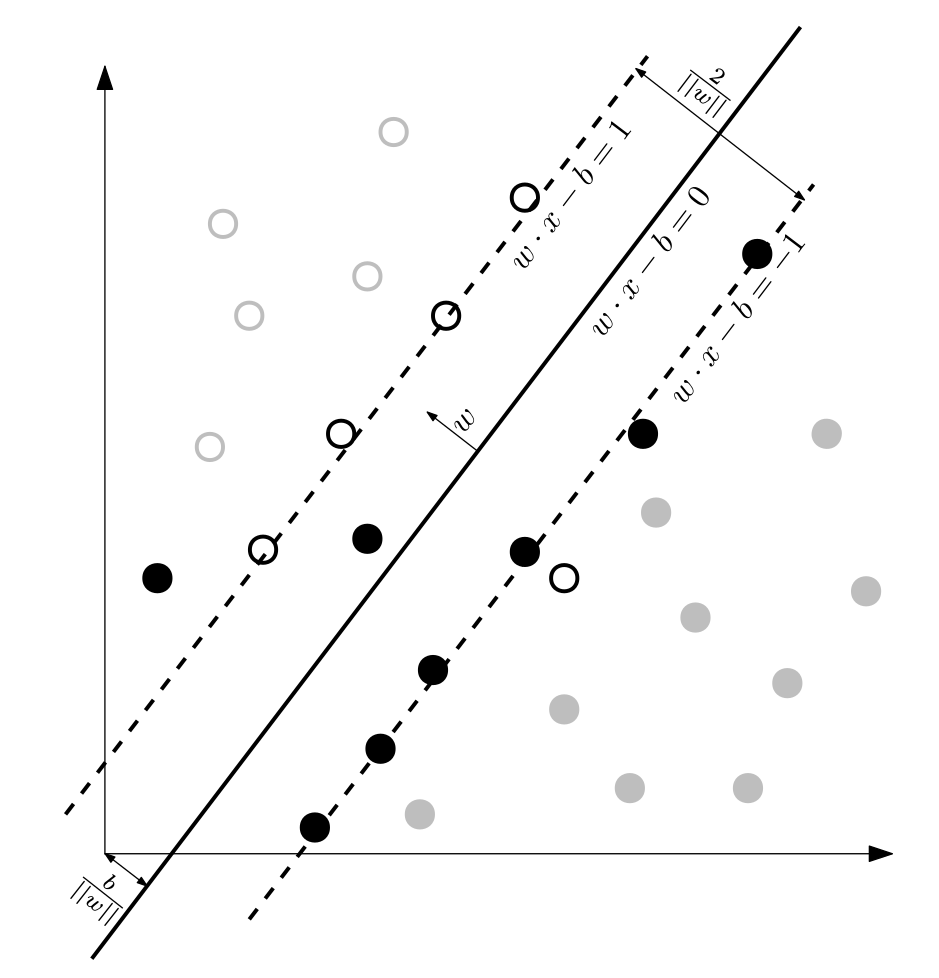}
  \caption{A non-separable binary classification problem solved with SVM and
    slack variables. Non-support vectors are gray shaded. The mapping $\phi$ is assumed to be the identity in this example.
\label{fig:svm_slack}}
\vspace*{-.25cm}
\end{figure}

A non-linearly separable example where slack variables are important and allow the SVM to find a separating hyperplane
is shown in Fig.~\ref{fig:svm_slack}.
We see that $\frac{2}{\lVert w \rVert}$ is the distance between the margins and
that in order to find the maximum separating hyperplane we need to minimize~$\lVert w \rVert$.
An easy-to-see but important consequence by the way the hyperplane is calculated
is that the maximum margin hyperplane is only determined by those $x_i$ which
lie nearest to it, \eg the data points whose removal would result in a
change of the hyperplane, known as  \emph{support vectors} (SVs).

\paragraph{Model Selection.}
Prior to training an SVM model, the model parameters have to be chosen.
In the case that is the main focus of the paper there are two parameters: the penalty parameter $C$ and the kernel parameter $\gamma$ of the Gaussian kernel. 
The optimization problem with fixed parameters is convex~\citep{cortes1995support}.
The model parameters are highly instance dependent.
Hence, parameter fitting is required to get optimal or near optimal parameters for a concrete
instance.
The parameter search is also called \emph{model selection}. This is a general problem not
only found in the context of SVMs.

\paragraph{Multi-class Classification.}
Our focus in this paper is on binary classification problems, but the concepts can
be extended to support multi-class classification where the $y_i$ are not
restricted to $\{+1, -1\}$ by allowing for multiple labels
directly~\citep{Crammer01} or training independent one-versus-rest binary SVMs \citep{IR-book}.

\section{Related Work}
\label{s:related}
There has been a \emph{huge} amount of research on machine learning and support vector machines so that we refer the reader to existing literature~\citep{Qiu2016,DBLP:journals/widm/Salcedo-SanzRMC14,DBLP:conf/birthday/StolpeBD16} for most of the material.
Here, we focus on issues closely related to our main contributions and previous work on the~problem. 

There have been many approaches that attempt to improve the performance of SVMs, \eg\citep{DBLP:journals/jmlr/FanCL05,joachims1998making,osuna1997improved}.
LibSVM~\citep{libsvm} is a widely used solver that implements the sequential minimal optimization algorithm.
LibLINEAR~\citep{DBLP:journals/jmlr/FanCHWL08} is a library that performs well on data where a non-linear SVM is not required, but is typically not feasible for complex data or data with a large amount of imbalance \citep{hsu2003practical}. 
Another attempt to improve the performance of SVMs is parallelization~\citep{DBLP:journals/siamjo/Mehrotra92,DBLP:conf/pdp/LiLDMW16,DBLP:conf/annpr/ToriiA06,DBLP:journals/jpdc/YouFSRKMYH15,DBLP:conf/nips/ChangZWBLQC07,cui2017multi,DBLP:conf/nips/GrafCBDV04,DBLP:conf/ipps/YouDCSV15}. In particular, there are methods that use parallel interior point methods to solve the underlying optimization problem~\citep{DBLP:journals/siamjo/Mehrotra92,DBLP:conf/nips/ChangZWBLQC07}. The approach by \cite{DBLP:conf/pdp/LiLDMW16} is also a parallel interior point method, but additionally uses GPUs. Other parallelizations use parallel stochastic gradient decent methods~\citep{DBLP:conf/icdm/ZhuCWZC09,platt199912} or try to avoid communication~\citep{cui2017multi,DBLP:conf/nips/GrafCBDV04,DBLP:conf/ipps/YouDCSV15}.

Another successful approach to improve the performance of SVMs are hierarchical techniques. 
The general multilevel paradigm originated from multigrid solvers for solving systems of
linear equations \citep{Sou35}. 
Multilevel schemes have been used in for a wide range of problems such as graph partitioning partitioning~\citep{kaffpa,karypis1998fast,DBLP:journals/corr/abs-2001-07134,DBLP:conf/gecco/MoreiraP018,DBLP:journals/siamsc/HerrmannOUKC19,DBLP:conf/wea/MoreiraPS17}, hypergraph partitioning~\citep{DBLP:conf/dac/KarypisK99,DBLP:conf/gecco/AndreS018}, graph clustering~\citep{DBLP:journals/corr/abs-0803-0476,DBLP:conf/aaim/DellingGSW09}, graph drawing~\citep{DBLP:journals/jgaa/Walshaw03,DBLP:journals/tvcg/MeyerhenkeN018} or the node separator problem~\citep{DBLP:conf/wea/Sanders016,DBLP:journals/coap/HagerHS18}.
In case of SVMs, there are currently two types of that scheme, i.e., early approaches that work in the feature space and later approaches that build a graph representation of the data first and then work with that representation to build a problem hierarchy.
The most general multilevel approach for SVMs consists of three main phases: training set coarsening, coarsest support vector learning, and support vector refinement.
In the \emph{contraction} (coarsening) phase, 
one iteratively decreases the problem size by performing contractions. Contraction should quickly reduce the size of the input and each computed level
should reflect the global structure of the input. 
Contraction is stopped when the problem is small enough so that a model can be trained by some other potentially more expensive algorithm. This is also called \emph{initial training} phase. 
In the \emph{refinement} (or uncoarsening) phase, contraction is iteratively undone, model parameters are inherited, and, at each level, a local improvement algorithm is
used to improve the quality of the result.  
In current multilevel SVMs the support vectors and optionally their neighbors
are uncontracted and used to train the SVM again using insights
from the parameter search of the~initial~training.

Yu \etal\citep{Yu03} were among the first to explore hierarchical SVM techniques. 
The authors create a hierarchically clustered representation of the data by merging data points based on distance. However, 
only linear classifiers have been considered and model parameters are neither inherited nor improved throughout the hierarchy.
Later, non-linear kernels have also been considered~\citep{DBLP:journals/eswa/HorngSCKCLP11} in an advanced intrusion detection system. Hsieh~\etal\citep{DC-SVM} also present an hierarchical approach that uses  a different geometric clustering~scheme.

Instead of using the feature space representation as previous works~\citep{Yu03,DBLP:journals/eswa/HorngSCKCLP11,DC-SVM}, \cite{razzaghi15} use a graph representation.
Their algorithm starts by building two graphs, one for each classification class, using approximate $k$-nearest neighbors.
The multilevel algorithm to train the SVM then uses this graph representation, i.e.,
the $k$-nearest neighbor graphs are gradually coarsened by performing contractions based on independent sets.
Moreover, the algorithm is the first to include \emph{refinement}, improving the initial trained model throughout the hierarchy from
coarsest to finest.
The support vectors of the previous coarser hierarchy level are used to train on the current level.
This approach is less sensitive to imbalanced data.
This is due to the fact that on the coarsest level the data classes are of roughly equal size because the
coarsening uses the same size-constraint for both classes individually.
This helps the initial SVM algorithm since the coarsest level is not imbalanced
anymore but still resembles the imbalanced original problem.
The approach is substantially faster with no loss of quality in the performance
measures, when compared the underlying SVM solver LibSVM~\citep{razzaghi15}.
\cite{amg_svm,amg_svm_arxiv} improved the result by
utilizing an algebraic multigrid (AMG) multilevel scheme, i.e.,
instead of independent set contractions an AMG algorithm is used in the coarsening
phase.

DC-SVM~\citep{DC-SVM} is a multilevel divide-and-conquer SVM that uses adaptive clustering.
It is one of the fastest SVMs, however, the approach by \cite{amg_svm,amg_svm_arxiv} demonstrates significantly better running time than DC-SVM on
almost all data sets~\citep{emlsvm}.

ThunderSVM~\citep{wenthundersvm18} is a parallel SVM library that runs on  GPUs as well as multi-core CPUs.
The algorithm reduces high-latency memory accesses and memory consumption through batch processing, kernel value reusing and sharing, and support vector sharing~\citep{thundersvm}.
For binary SVMs, kernel values are computed in batches with the possibility of
reusing them later via a buffer.
On the other hand, LibSVM provides a simple modification which
  parallelizes the library trivially using OpenMP~\citep{libsvm_faq}. 
  More precisely, using OpenMP, LibSVM uses parallel reductions to parallelize kernel evaluations in training/testing.
The multi-threaded CPU version of ThunderSVM outperforms the parallelized LibSVM by a
factor of 2-4 on binary classification and up to 10 on multi-class
classification. When comparing the CPU version to the single threaded LibSVM
version ThunderSVM reaches a speed up of 50~\citep{thundersvm}.

\section{Scalable SVMs by Cluster Contraction}
We now give a full description of our algorithm. Note that our algorithm is similar to the one of Razzaghi and Safro~\citep{razzaghi15}, i.e.,
we transfer the problem to a graph problem and then use a graph-based multilevel algorithm to build a hierarchy of (smaller) training problems. 
Our algorithm, however, uses a much simpler and more efficient coarsening strategy that is able to shrink the problem size very quickly and thus is much faster while giving similar classification results. 

\subsection{Overview}
An overview of our approach is shown in Algorithm~\ref{algo:overview}.
Our algorithm starts by preprocessing the data and afterwards splitting it into the two given classes
$\mathbf{C}^+$ and $\mathbf{C}^-$.
We then  build \emph{two} $k$-nearest neighbor graphs, one for each class. The rest of the algorithm is a multilevel approach that works on those two graphs.
Our algorithm starts coarsening by computing coarse versions of each of the two graphs independently using a \emph{clustering contraction} scheme in each graph. 
More precisely, we use and compare two different types of algorithms to obtain clusterings. First,  we use a label propagation algorithm to compute a clustering and secondly, we use an algorithm that computes clusterings in which clusters have a small diameter. After computing a clustering with one of those algorithms, we contract the clusters. We continue to do this recursively.
For each of the two graphs, coarsening is stopped once the associated graph has a predefined size. 
Note that since the coarsening process is done independently on each of the graphs, the coarsest graphs are roughly balanced in the number of vertices they contain.
When coarsening has been stopped on both graphs, we perform \emph{initial training} using the coarsest level of the computed hierarchies, i.e., we solve the SVM optimization problem on the coarsest graphs of that hierarchy.
Once the initial model is trained, we recursively uncontract the next finer
hierarchy level and train a new SVM model on the support vectors of the previous model.
After the finest level is processed and we trained models for every level of
the hierarchy, we choose the overall best model as the final model of the
complete multilevel process. We now give full details.

\subsection{Preprocessing Data}
Our algorithm starts with two commonly used preprocessing routines: categorical feature preprocessing and feature scaling.

\paragraph{Categorical Features.}
SVMs require data to be represented by real valued numbers.
Often times features such as the country someone lives in or the marital status
are given as categorical variables.
Categorical means that the feature can take one of a limited and fixed number of
possible values.
For categorical attributes an extra preprocessing step is needed to convert them
into numerical data.
We do so by using one-hot encoding~\citep{Harris13}.
An $\ell$-category attribute is represented by $\ell$ numbers where one of the
$\ell$ numbers is one and the others are zero.
If the number of values in an attribute is not too large,
this coding is more stable than using a single number~\citep{Hsu10}.

\paragraph{Scaling.}
The purpose of scaling is to avoid that attributes in greater numeric
ranges dominate those in smaller numeric ranges~\citep{Hsu10}.
The problem can be avoided by scaling the feature columns
independently.
Our algorithm uses the standardization technique~\citep{StolckeKF08}.
Here, each of the feature columns is transformed by
subtracting the mean and then dividing by the standard deviation such that all
feature columns have zero-mean and unit-variance.

\subsection{Graph Model}
We transform the training problem into a graph problem  using the technique of Razzaghi and Safro~\citep{razzaghi15}.
More precisely, we compute two graphs, one for each of the classes $\mathbf{C}^+$ and $\mathbf{C}^-$.
For a class, the graph is constructed by applying an approximate $k$-nearest neighbor algorithm on the data.
Thus, we obtain a graph with information about the \emph{proximity} of every vertex.
We use the A$k$NN library FLANN~\citep{flann} to compute the $k$-nearest neighbor graph.
Afterwards we set the weight of an edge to $1/\text{dist}(p,q)$ for two points that are adjacent in the graph, where $\text{dist}(\cdot,\cdot)$ is the Euclidean distance.
This is done since the clustering algorithms used for coarsening intrinsically try find clusters that are internally dense (high-edge weights) and externally spare (low edge-weights).
Hence, by using $1/\text{dist}(p,q)$ as edge weights, points that are close are more likely to be clustered and later on contracted. 
\paragraph{Contraction}
Once we have computed a clustering, \emph{contracting it} works as follows: 
each block of the clustering is contracted into a single node. 
Note that each node has an associated feature vector.
The feature vector of the coarse representative is computed as follows:
for each feature column, the value of the feature of the coarse representative is set to the arithmetic mean of the values of the feature of all nodes that are contracted into that node.
There is an edge between two nodes $u$ and $v$ in the contracted graph if the
two corresponding blocks in the clustering are adjacent to each other in $G$,
i.e., block $u$ and block $v$ are connected by at least one edge.
The weight of an edge $(A,B)$ is set to inverse distance of the corresponding feature vectors.

\begin{algorithm}[t]
 preprocess data\;

   build $k$-nearest neighbor graph for $C^+$ and $C^-$\; \\
   contract graphs recursively, build hierarchy\; \\
   initial training on coarsest problem\; \\
   \While{levels in the hierarchy}{
     train SVM model on uncontracted support vectors of previous level\;
   }
   \Return best model of all levels\;
 \caption{Overview}
 \label{algo:overview}
\end{algorithm}

\subsection{Coarsening by Cluster-Contraction}

Our algorithm computes coarse versions of each of the two graphs independently
using a \emph{clustering contraction}~scheme.
To compute a graph hierarchy, the clustering is contracted by replacing each cluster/block by a single node, and the process is repeated recursively until the graph is small.
This way the inherent cluster hierarchy of the networks is detected and the contraction of important edges in small cuts is unlikely.
For the original problem this means that we contract groups of data points that are close to each other into one single data point.
Note that cluster contraction is an aggressive coarsening strategy. In contrast to most previous approaches, it can drastically shrink the size of irregular networks.
Experiments by \cite{pcomplexnetworksviacluster} indicate that already one contraction step can shrink the graph size by orders of magnitude. 
We consider two different clustering schemes:

\paragraph{Label Propagation.}
We use a variation of the \emph{label propagation} algorithm proposed by
\cite{labelpropagationclustering}, which is a very fast, near
linear-time algorithm that locally optimizes the number of edges cut.
Initially, each node is in its own cluster/block, i.e., the initial block ID of
a node is set to its node ID.
The algorithm then works in rounds. 

In each round, the original algorithm traverses the nodes of the graph in random order.
However, previous work~\citep{pcomplexnetworksviacluster} has shown that using the ordering induced by the node degree (increasing order) improves the overall solution quality \emph{and} running time.
Using this node ordering means that in the first round of the label propagation algorithm, nodes with small node degree can change their cluster before nodes with a large node degree. We use this node ordering in our algorithm.

When a node $v$ is visited, it is \emph{moved} to the cluster that has the strongest connection to $v$, i.e., it is moved to the cluster $V_i$ that maximizes $\omega(\{(v, u) \mid u \in N(v) \cap V_i \})$. 
Ties are broken randomly. 
Originally, the process is repeated until it has converged. 
We perform at most $\ell$ rounds of the algorithm instead, where $\ell$ is a tuning parameter.
One round can be implemented to run in linear time. 
Note that due to the way weights are defined in our graphs, i.e., edge weights are anti-proportional to the distance of vertices, our algorithm finds clusters of vertices that are \emph{close} to each other.
We also tried a variant of the label propagation algorithm that, in contrast to the original algorithm~\citep{labelpropagationclustering}, 
ensures that each block of the clustering fulfills a size constraint. However, we observed that this slows down the overall algorithm and had almost no effect on the observed~classification~quality.

\paragraph{Low Diameter Clustering.}
A low diameter clustering partitions the vertices such that the
diameter of each cluster is small and the number of edges between clusters
is small. 
We are interested in small diameter clusterings for contraction since they intuitively yield less support vectors on each level of the multilevel hierarchy and lead to a better approximate of the original problem.  
Hence, intuitively using small diameter clusterings improves quality and speed of the overall algorithm.
\cite{miller13} propose a linear-work algorithm to compute low-diameter decompositions.
Here, we use a simpler and faster implementation of the algorithm by \cite{shun14}. 
\cite{miller13} give a strong bound for the maximum
diameter of the clustering $O(\log n/\beta)$ where $n$ is the number of
vertices.
Note that the $\beta$ parameter allows for defining the maximum diameter of the
clusters. This gives us a parameter to control the depth of the coarsening
hierarchy since smaller diameters result in more clusters and ultimately more
levels in the hierarchy. 

Intuitively, the algorithm can be viewed as performing breadth-first searches
from different starting vertices in parallel with start times drawn from an
exponential distribution~\citep{shun14}.
We follow their description closely: The algorithm starts by assigning each vertex $v$ a shift value $\delta_{v}$
 drawn from an exponential distribution with parameter $\beta$ (mean
 $1/\beta$).
 Each vertex $v$ is then assigned to the cluster $S_u$ that minimizes the shifted distance $\text{dist}_{-\delta}(u,v) = \text{dist}(u,v) - \delta_u$.
 This can be implemented by performing multiple BFS's in parallel.
 Each iteration of the implementation explores one level of each BFS and at
 iteration $t$ (starting with $t = 0$) breadth-first searches are started from the
 unvisited vertices $v$ such that $\delta_{v} \in [t, t + 1)$.
 If a vertex is visited by multiple BFSs in the same iteration the tie is broken
 by comparing the smaller fractional portion of the shift value.

\subsection{Initial Training}
As soon as the graphs are small enough, we train the initial model.
We perform \emph{model selection} and solve the SVM optimization problems by using the C-SVM routine of the
underlying SVM library -- which, depending on the configuration of our algorithm, is either LibSVM or ThunderSVM.
As our algorithm uses the Gaussian kernel to allow for non-linear classification, we have to find good model parameters $C$~and~$\gamma$.

A common way to find the best parameters in SVM learning is to use grid
search~\citep{svm_nested_grids} over the parameter space where a regular grid is put on the parameter space and
the parameter combinations corresponding to the grid vertices are evaluated.
However, while grid search is robust it is 
work intensive.
In contrast to grid search, Uniform Design (UD)~\citep{UD} uses a
pre-computed set of parameters that have maximal distance to each other.
This ensures that the parameter space is covered well but reduces the number of parameters that have to be evaluated.
Our implementation of UD is similar to the one used~by~\cite{razzaghi15} and performs a two sweep search.
That is we begin the search using points that are well distributed and in a second sweep we use points close to the best parameters found during the first sweep.
Following \cite{razzaghi15} the first sweep evaluates 9 parameter combinations and the second one~4~combinations.

\subsection{Uncoarsening}
After initial training, our algorithm improves the solution on every level of the hierarchy.
We uncontract the hierarchies for the majority and
minority class at the same time.
Recall that support vectors are the 
important data points for the construction of the maximum separating hyperplane.
We use this property to decrease the problem~size.

Consider an SVM model trained on the previous, coarser level.
The support vectors of the model are data points of the previous levels' problem.
For the current level, we uncontract those support vectors and use the resulting
data points as input to train a new SVM model. Solving the optimization problem is again done using the SVM library.
In the case that there are more hierarchy levels for the majority class than for
the minority class, which is often the case for imbalanced data, we only
uncontract the majority hierarchy until both hierarchies have similar~size.

Our algorithm also adopts the model parameters from the previous level. 
The assumption is that model parameters of coarser levels
 are already good model parameters for the current level.
We only perform a second, more fine-grained UD sweep around the model parameters of the previous level to improve the model.
When the problem size exceeds \numprint{10000} data points, we adopt the best parameters from the previous level and skip the model selection process.
We proceed uncontraction recursively until the complete hierarchy
is processed.
We return the best model that we found during uncoarsening.
Note that due to overfitting this does not have to be the SVM model of the last level.

\subsection{Evaluation of Solutions}
At different stages of the algorithm we need to assess the quality of the current solution.
This is usually done by using data points with known labels and comparing the known correct label to the
label which the trained SVM predicts for the data point.
More precisely, in machine learning it is common to split a given instance into two parts: the
training set and the test set~\citep{IML}.
Learning is conducted on the training set and evaluation is done on the test set.
However, we can not use test data for the evaluation tasks during the training phase, otherwise the result of the algorithm would not be meaningful.

Sadrfaridpour~\etal\citep{emlsvm} discuss several different evaluation approaches.
We use the approach that worked best in their studies:
Rather than using the data of the current level that our algorithm works on, we use the original problem (i.e., the training
set) as validation set in our algorithm. Since we aim at classifying large problems we can
not validate every parameter combination on the entire training set, which is why
we use a random subset of the training set for evaluation. The size of the validation set is 10\% of the original training data.

\subsection{Shared-Memory Parallelization}
Our parallelization has a focus on the training phase as repeated SVM training that is done during the uncoarsening phase is the most time consuming part of our multilevel approach.
Other parts of the overall algorithm could potentially be parallelized, too -- such as coarsening.
However, already on medium sized graphs coarsening consumes less than 5\% of the algorithm's overall running time.
Moreover, the relative running time of coarsening decreases even more with increasing instance size so that the effort does not seem worth it.
In our parallel algorithm, we replace LibSVM -- the library used to train the models
during uncoarsening --  with ThunderSVM~\citep{wenthundersvm18}.

\section{Experimental Evaluation}\label{sec:eval}

\subsection*{Systems and Methodology}\label{Methodology}
We implemented the algorithm described in the previous section using C++. 
Our code uses FLANN~1.8.4~\citep{flann},
LibSVM~3.22~\citep{libsvm}, and is compiled with gcc-8.1.1.
Sequential experiments are executed on a single core machine with four AMD Opteron 6168 (12 cores) with 1.9GHz and 256GB of RAM.
For parallel experiments, we use all cores of that machine.
Due to the large running times of both LibSVM and ThunderSVM, we use a time limit of 24 hours in our experiments. Instance timeouts are reported as a dash.
The rest of this section is structured as follows. First, we outline instances, performance measures, and algorithm configurations.
  We then compare different algorithms and configurations of our algorithm with enabled model selection, i.e., all of the algorithms also optimize the \emph{model parameters} $C$ and $\gamma$ for their
  final output (see Section~\ref{sec:withmodelparameters}). Afterwards, we use the largest instances available to us and run our parallel algorithms for a single set of parameters $C$, $\gamma$ taken from previous work (see Section~\ref{sec:largeinstances}).

\begin{table}[b!]
  \centering
  \scriptsize
  \begin{tabular}{l|rrrrr}
    Name & Size & Feat. & $|\mathbf{C}^+|$ & $|\mathbf{C}^-|$ & Imb.\\
    \hline
    Advertisement    & 3\ 279   & 1\ 558  & 459      & 2\ 820   & 0.86 \\
    APS failure      & 76\ 000  & 170 & 1\ 375  & 74\ 625  & 0.98 \\
    Buzz             & 140\ 707 & 77      & 27\ 775  & 112\ 932 & 0.80 \\
    Census           & 299\ 285 & 41  & 18\ 568 & 280\ 717 & 0.94 \\
    Clean (Musk)     & 6\ 598   & 166     & 1\ 017   & 5\ 581   & 0.85 \\
    Cod-rna          & 59\ 535  & 8       & 19\ 845  & 39\ 690  & 0.67 \\
    EEG Eye State    & 14\ 980  & 14      & 6\ 723   & 8\ 257   & 0.55 \\
    Forest (Class 3) & 581\ 012 & 54      & 35\ 754  & 545\ 258 & 0.94 \\
    Forest (Class 5) & 581\ 012 & 54      & 9\ 493   & 571\ 519 & 0.98 \\
    Forest (Class 7) & 581\ 012 & 54      & 20\ 510  & 560\ 502 & 0.96 \\
    Hypothyroid      & 3\ 919   & 21      & 240      & 3\ 679   & 0.94 \\
    Isolet (Class A) & 6\ 919   & 617     & 240      & 5\ 998   & 0.96 \\
    Letter (Class Z) & 20\ 000  & 16      & 734      & 19\ 266  & 0.96 \\
    Letter (Class A) & 20\ 000  & 16  & 786     & 19\ 266  & 0.96 \\
    Letter (Class B) & 20\ 000  & 16  & 766     & 19\ 266  & 0.96 \\
    Letter (Class H) & 20\ 000  & 16  & 734     & 19\ 266  & 0.96 \\

    Nursery          & 12\ 960  & 8       & 4\ 320   & 8\ 640   & 0.67 \\
    Protein          & 145\ 751 & 74      & 1\ 296   & 144\ 455 & 0.99 \\
    Ringnorm         & 7\ 400   & 20      & 3\ 664   & 3\ 736   & 0.50 \\
    Skin             & 245\ 057 & 3   & 50\ 859 & 194\ 198 & 0.79 \\
    Sleep (Class 1)  & 105\ 908 & 13  & 9\ 052  & 96\ 856  & 0.91 \\
    Twonorm          & 7\ 400   & 20      & 3\ 703   & 3\ 697   & 0.50 \\
    \hline
    
    \hline
    Mnist8m (Class 0) & 8\ 100\ 000 & 784   & 799\ 605    & 7\ 300\ 395 & 0.90 \\
    Mnist8m (Class 1) & 8\ 100\ 000 & 784   & 910\ 170    & 7\ 189\ 830 & 0.89 \\
    Susy              & 5\ 000\ 000 & 18    & 2\ 287\ 827 & 2\ 712\ 173 & 0.54 \\

  \end{tabular}
  \caption{Properties of instances in our benchmark set.}
  \label{tab:data}
  \vspace*{-.5cm}
\end{table}

\subsection*{Instances}
We use a super set of instances  that are used by \cite{emlsvm}, i.e., we use the same instances and extend the benchmark set with larger problem instances.
Most of the instances are from the UC Irvine Machine Learning Repository~\citep{UCI}. The one exception is Mnist8m which is from \citep{loosli2007training}.
Basic properties of the instances in our benchmark set are shown in \autoref{tab:data}. Here, size refers to the number of labeled inputs, features (feat.)~refers to the number features, $|\mathbf{C}^+|$ to the number of inputs having a positive label, $|\mathbf{C}^-|$ to the number of inputs from the majority class and imbalance (Imb.) refers to $\max\{|\mathbf{C}^+|, |\mathbf{C}^-|\}/(|\mathbf{C}^+|+|\mathbf{C}^-|)$.

\subsection*{$k$-Fold Cross-Validation}
In order to evaluate the performance of the algorithms, a given instance is split into the training set,
which is the only data used during training, and the test set.
After solving the SVM optimization problem for the training set, one then compares labels of the \emph{test} data computed by the final SVM with the given labels to measure prediction~quality. 
We use $k$-fold cross-validation~\citep{DBLP:conf/ijcai/Kohavi95} to get a more accurate estimate
of the prediction performance.
More precisely, we first shuffle the entire data set and split it into $k$
parts of equal size.
We then perform $k$ repetitions of the training algorithm.
In every run one of the parts is used as the \textit{test set} while the
other $k-1$ parts constitute the \textit{training set}.
The test set is \emph{never} used to train or validate any of the intermediate results
of the hierarchy; only the final result of a run is evaluated using the
test set.
After all $k$ runs are finished, we average the results of all runs and present
that as the overall quality of our training process. 
We use $k=5$ in our experiments, since this is the default value in mlsvm-AMG~\citep{emlsvm}.
In our experiments, all algorithms operate on the same $k$-folds in order to achieve~meaningful~comparisons.
By default we perform five $k$-folds for each algorithm using different random seeds. Hence, in total we perform 25 different runs for
each test instance and report the arithmetic mean of solution quality and running time.
When further averaging over multiple instances, we use the geometric mean in
order to give every instance a comparable influence on the final~score.

\subsection*{Algorithm Configuration}
Any multilevel algorithm has a considerable number of choices between
algorithmic components and tuning parameters. 
The model selection parameters $C$ and $\gamma$ are not part of the algorithm configuration as they are
determined during the initial SVM training.
The number $\ell$ of label propagation iterations during coarsening is fixed to ten, as more iterations rarely found better clusterings for contraction in previous studies such as~\citep{DBLP:journals/tpds/MeyerhenkeSS17}.
We stop the coarsening process, when the graph size is smaller than 500, as this is the default value in \citep{razzaghi15}. 
Our algorithm uses the same value $k=10$ for the construction of the $k$-nearest neighbor graphs as mlsvm-AMG~\citep{emlsvm}. 
This configuration of our algorithm is called LPSVM or quality configuration unless otherwise mentioned.
We also use a fast configuration of our algorithm, LPSVM$_\text{fast}$. This configuration returns the model that has been found after the initial training phase and does not perform any further refinement during uncoarsening.
Based on preliminary experiments on a different set of instances used for evaluation here, our LowDia algorithms use the low diameter coarsening scheme with $\beta := 0.4$.
If we use ThunderSVM as training algorithm during uncoarsening the corresponding configuration is marked with $_\text{par}$.
All parallel experiments are run on 32 cores.
In Section~\ref{sec:largeinstances}, we also employ a random sampling algorithm to preprocess the training instances. Our algorithm uses $p*|I|$ random samples  from the training instance -- here, $|I|$ refers to the size of the training instance. We mark an algorithm configuration using this the sampling technique by writing $p$ as a subscript next to the configuration name.

\subsection*{Performance Measures}
\label{ssub:performance_measures}

We employ the commonly used performance measures \textit{sensitivity} (SN), \textit{specificity}
(SP), and \textit{G-mean}, 
to evaluate our prediction results \citep{emlsvm}.
The definition for those are
\begin{align*}
  \label{eq:measures}
  SN &= \frac{TP}{TP + FN}\\
  SP &= \frac{TN}{TN + FP}\\
  G\text{-mean} &= \sqrt{SP \cdot SN},
\end{align*}
where TP are the true positives (the correctly classified points of
$\mathbf{C}^+$), FN the false negatives (wrongly classified points of the
minority class $\mathbf{C}^+$), TN the true negatives (correctly classified point of
$\mathbf{C}^-$), and FP the false positives (points of $\mathbf{{C}^-}$
wrongly classified as points of the minority class).
These metrics are common in statistical analysis with accuracy being the most
used metric in machine learning~\citep{IML}.
Note that on large imbalanced data sets getting high
accuracy is trivial by only predicting the larger class.
Hence, as we work with imbalanced instances, we cannot use accuracy as primary measure for prediction quality.
Instead, we use the geometric mean of the sensitivity and the specificity.
 \cite{mlsvm_healthcare} use this metric because it is very sensitive to false
negatives. In general, this metric yields more informative results on imbalanced data sets.  
Moreover, when comparing trained SVM models, \eg in model selection during 
training or later when searching for the best overall SVM model, we use the
G-mean as primary decision criterion. If two models have roughly the same G-mean
we choose the model with less support~vectors since this is likely to be
a better abstraction.

\subsection{Comparison with the State of the Art}
\label{sec:withmodelparameters}
\begin{table*}[t]
  \scriptsize
  \centering
  \begin{tabular}{l||cccccc|cc}
    {\multirow{2}{*}{}} & mlsvm & LP  & LP                & LP               & Low & Low              & Lib & Thunder \\
    {\multirow{2}{*}{}} & -AMG  & SVM & SVM$_\text{fast}$ & SVM$_\text{par}$ & Dia & Dia$_\text{par}$ & SVM & SVM     \\

        Instance &  \multicolumn{6}{c|}{G-mean} & \multicolumn{2}{c}{G-mean}\\
    
    \hline
    \hline
\texttt{Advertisement}  & 0.91 & 0.84 & 0.75 & 0.78 & 0.82 & 0.83 & 0.94         & 0.92         \\
\texttt{APS failure}    & 0.94 & 0.94 & 0.93 & 0.93 & 0.94 & 0.94 & -            & 0.83         \\
\texttt{Buzz}           & 0.95 & 0.92 & 0.92 & 0.92 & 0.90 & 0.95 & -            & 0.94         \\
\texttt{Census}         & 0.81 & 0.81 & 0.80 & 0.81 & 0.81 & 0.81 & -            & -            \\
\texttt{Clean (Musk)}   & 0.88 & 0.94 & 0.87 & 0.92 & 0.96 & 0.97 & 1.00         & 0.98         \\
\texttt{Cod-rna}        & 0.94 & 0.94 & 0.94 & 0.94 & 0.94 & 0.94 & 0.86         & 0.96         \\
\texttt{EEG Eye State}  & 0.75 & 0.75 & 0.54 & 0.72 & 0.76 & 0.75 & 0.93         & 0.87         \\
\texttt{Forest (Cl. 3)} & 0.94 & 0.91 & 0.92 & 0.92 & 0.94 & 0.92 & -            & -            \\
\texttt{Forest (Cl. 5)} & 0.79 & 0.84 & 0.75 & 0.85 & 0.86 & 0.86 & -            & 0.92         \\
\texttt{Forest (Cl. 7)} & 0.86 & 0.88 & 0.86 & 0.90 & 0.93 & 0.93 & -            & -            \\
\texttt{Hypothyroid}    & 0.94 & 0.84 & 0.87 & 0.83 & 0.90 & 0.90 & 0.93         & 0.94         \\
\texttt{Isolet (Cl. A)} & 0.00 & 0.99 & 0.89 & 0.99 & 0.95 & 0.95 & 0.99         & 0.98         \\
\texttt{Letter (Cl. A)} & 0.94 & 0.96 & 0.95 & 0.96 & 0.97 & 0.98 & 0.99         & 0.99         \\
\texttt{Letter (Cl. B)} & 0.88 & 0.93 & 0.91 & 0.93 & 0.95 & 0.95 & 0.98         & 0.96         \\
\texttt{Letter (Cl. H)} & 0.76 & 0.90 & 0.85 & 0.90 & 0.92 & 0.92 & 0.97         & 0.93         \\
\texttt{Letter (Cl. Z)} & 0.92 & 0.95 & 0.95 & 0.96 & 0.96 & 0.96 & 0.99         & 0.99         \\
\texttt{Nursery}        & 1.00 & 1.00 & 1.00 & 1.00 & 1.00 & 1.00 & 1.00         & 1.00         \\
\texttt{Protein}        & 0.92 & 0.93 & 0.93 & 0.93 & 0.92 & 0.92 & -            & 0.88         \\
\texttt{Ringnorm}       & 0.98 & 0.97 & 0.85 & 0.97 & 0.95 & 0.95 & 0.99         & 0.97         \\
\texttt{Skin}           & 0.99 & 0.98 & 0.99 & 0.98 & 0.97 & 0.98 & 1.00         & 1.00         \\
\texttt{Sleep (Cl. 1)}  & 0.69 & 0.68 & 0.69 & 0.66 & 0.57 & 0.56 & -            & 0.50         \\
\texttt{Twonorm}        & 0.97 & 0.97 & 0.97 & 0.96 & 0.97 & 0.97 & 0.98         & 0.97         \\
        \hline
        avgerage score  & 0.85 & 0.90 & 0.87 & 0.90 & 0.91 & 0.91 & 0.97* (0.66) & 0.92* (0.80) \\

  \end{tabular}
  \caption{Computational results of different algorithms to train an SVM including model parameter selection $C$ and $\gamma$. Timeouts are reported as dash '-'. Avg. scores marked with a '*' are computed only on the instances that the algorithm could solve -- (x) shows the score if we assume a G-mean score of 0 in case that an algorithm is not able to solve an~instance.}
  \label{tab:comp}
\end{table*}

\begin{table*}[t]
  \scriptsize
  \centering
  \begin{tabular}{l||rrrrrr|rr}
    {\multirow{2}{*}{}} & \multicolumn{1}{c}{mlsvm} & \multicolumn{1}{c}{LP} & \multicolumn{1}{c}{LP} & \multicolumn{1}{c}{LP} & \multicolumn{1}{c}{Low} & \multicolumn{1}{c}{Low} & \multicolumn{1}{|c}{Lib} & \multicolumn{1}{c}{Thunder}\\
    {\multirow{2}{*}{}} & \multicolumn{1}{c}{-AMG}  & \multicolumn{1}{c}{SVM} & \multicolumn{1}{c}{SVM$_\text{fast}$} & \multicolumn{1}{c}{SVM$_\text{par}$} & \multicolumn{1}{c}{Dia} & \multicolumn{1}{c}{Dia$_\text{par}$} & \multicolumn{1}{|c}{SVM} & \multicolumn{1}{c}{SVM}\\
    Instance &  \multicolumn{6}{c|}{Running time [s]} & \multicolumn{2}{c}{Running time [s]}\\
    
    \hline
    \hline
\texttt{Advertisement}    & 343     & 145    & 31  & 184    & 262    & 155    & 557     & 234     \\
\texttt{APS failure}      & 1\ 473  & 88     & 24  & 31     & 215    & 56     & -       & 3\ 235  \\
\texttt{Buzz}             & 110     & 264    & 16  & 414    & 365    & 272    & -       & 39\ 026 \\
\texttt{Census}           & 3\ 047  & 240    & 38  & 163    & 622    & 564    & -       & -       \\
\texttt{Clean (Musk)}     & 14      & 8      & 4   & 5      & 19     & 10     & 320     & 29      \\
\texttt{Cod-rna}          & 80      & 24     & 4   & 43     & 11     & 14     & 15\ 700 & 1\ 640  \\
\texttt{EEG Eye State}    & 123     & 864    & 0.6 & 1\ 583 & 496    & 631    & 2\ 700  & 1\ 157  \\
\texttt{Forest (Class 3)} & 10\ 156 & 718    & 81  & 437    & 883    & 704    & -       & -       \\
\texttt{Forest (Class 5)} & 6\ 986  & 648    & 60  & 364    & 787    & 358    & -       & 68\ 579 \\
\texttt{Forest (Class 7)} & 5\ 393  & 300    & 78  & 175    & 382    & 209    & -       & -       \\
\texttt{Hypothyroid}      & 2       & 3      & 0.7 & 4      & 1      & 2      & 13      & 2       \\
\texttt{Isolet (Class A)} & 1\ 627  & 25     & 8   & 9      & 37     & 12     & 856     & 217     \\
\texttt{Letter (Class A)} & 17      & 3      & 2   & 2      & 5      & 4      & 1\ 930  & 100     \\
\texttt{Letter (Class B)} & 55      & 4      & 2   & 4      & 7      & 6      & 1\ 590  & 105     \\
\texttt{Letter (Class H)} & 74      & 8      & 2   & 13     & 13     & 15     & 1\ 970  & 124     \\
\texttt{Letter (Class Z)} & 31      & 3      & 2   & 2      & 5      & 3      & 1\ 710  & 97      \\
\texttt{Nursery}          & 7       & 2      & 0.6 & 2      & 4      & 4      & 998     & 45      \\
\texttt{Protein}          & 3\ 654  & 30     & 14  & 28     & 148    & 74     & -       & 3\ 205  \\
\texttt{Ringnorm}         & 10      & 14     & 0.5 & 4      & 4      & 7      & 161     & 24      \\
\texttt{Skin}             & 81      & 16     & 10  & 10     & 31     & 18     & 38\ 200 & 4\ 210  \\
\texttt{Sleep (Class 1)}  & 1\ 594  & 1\ 450 & 7.4 & 3\ 357 & 1\ 344 & 2\ 944 & -       & 35\ 606 \\
\texttt{Twonorm}          & 7       & 0.6    & 0.4 & 0.7    & 2.1    & 2.5    & 109     & 15      \\

  \end{tabular}
  \caption{Running times of different algorithms to train an SVM including model parameter selection $C$ and $\gamma$. Timeouts are reported as dash '-'. Running times are rounded.}
  \label{tab:comprunningtimes}
\end{table*}

\begin{figure}[t]
\begin{center}
\includegraphics[width=.32\textwidth]{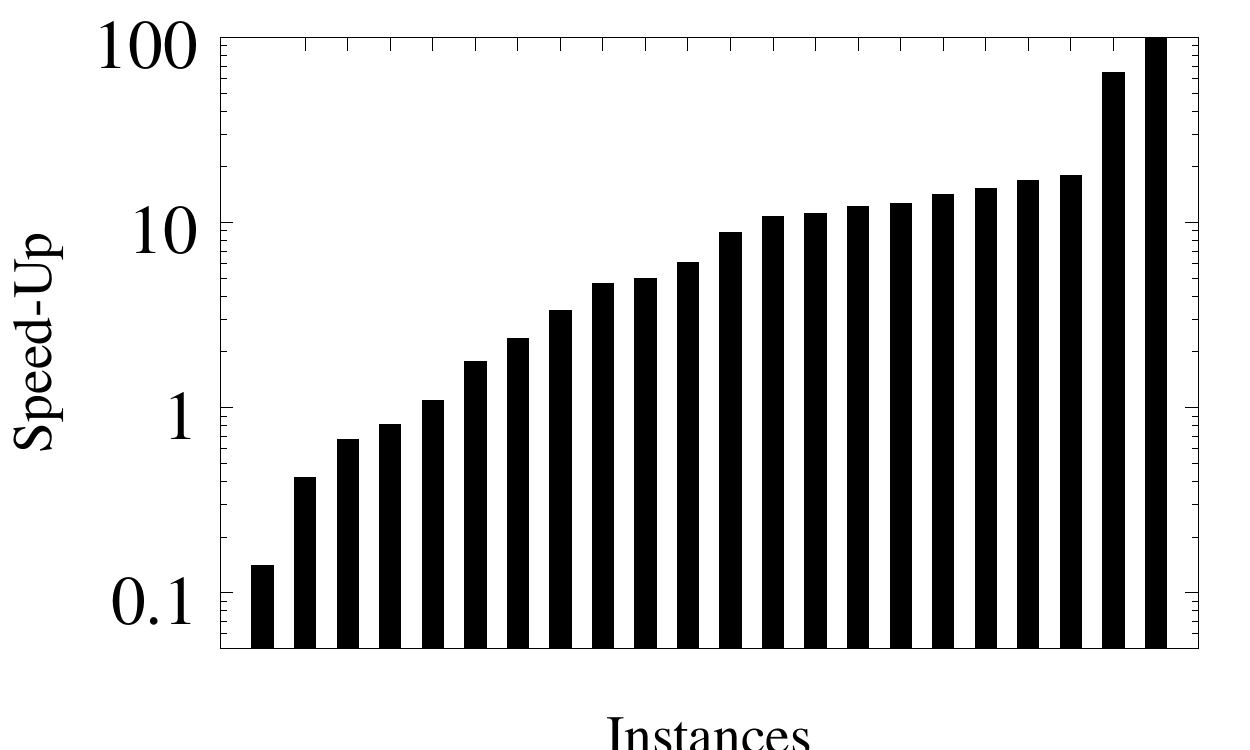}
\includegraphics[width=.32\textwidth]{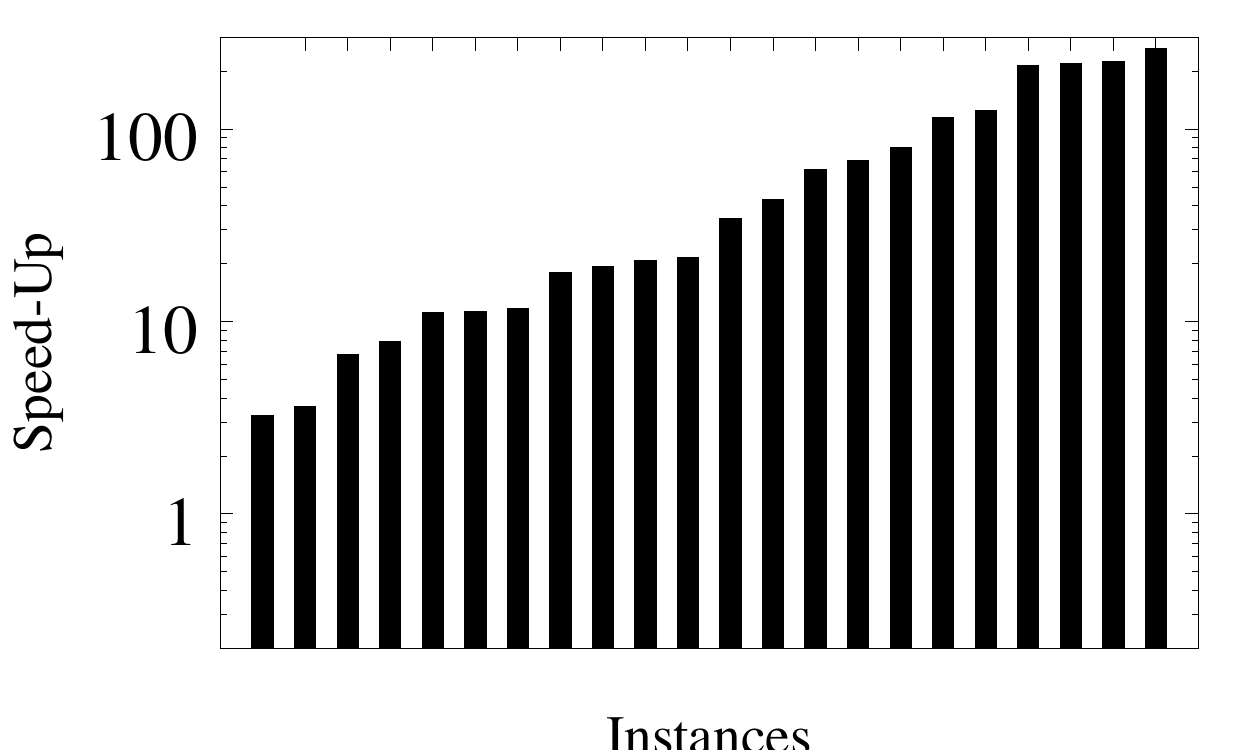}
\includegraphics[width=.32\textwidth]{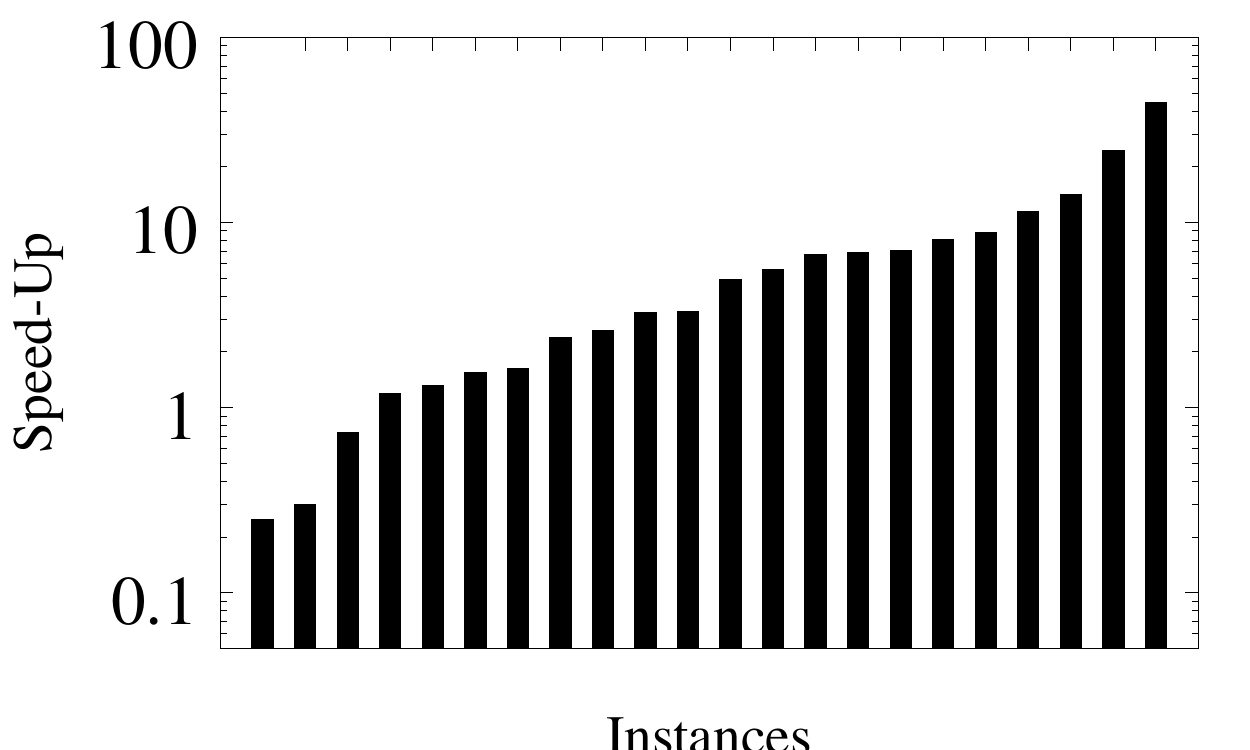}
\end{center}
\caption{Distribution of speed-ups of our algorithm over mlsvm-AMG when running
  algorithms on the same $k$-folds. From left to right: speed-ups of LPSVM over
  mlsvm-AMG. speed-ups of LPSVM$_\text{fast}$ over mlsvm-AMG, speed-ups of LowDia over mlsvm-AMG.}
\label{fig:distspeedups}
\vspace*{-.5cm}
\end{figure}

In this section, we compare different algorithms and configurations of our algorithm with enabled model selection, i.e.,
  the optimization process of all of the algorithms also \emph{finds good model parameters} $C$ and $\gamma$.
We compare our algorithm to mlsvm-AMG (the
previously best sequential system), and to the solvers LibSVM and ThunderSVM.
In any case, we stopped computations on an instance if the running time exceeded 24 hours.
We do not perform additional comparisons with mlsvm-IS~\citep{razzaghi15}, DC-SVM~\citep{DC-SVM}, and EnsembleSVM~\citep{DBLP:journals/jmlr/ClaesenSSM14} as mlsvm-AMG computes similar or improved classification results while being up to one to two orders of magnitude faster on large instances in case of the first three algorithms, or in case of EnsembleSVM~\citep{DBLP:journals/jmlr/ClaesenSSM14} the reported classification quality of mlsvm-AMG is significantly higher.
We run the algorithms on our machine without tuning parameters on a per-instance basis, i.e, we use the same set of parameters as described above for every instance. In the original work, mlsvm-AMG~\citep{amg_svm,amg_svm_arxiv} uses instance-based parameters, i.e.,   different instances use different parameters of the multilevel algorithm.\footnote{Personal communication with Ehsan Sadrfaridpour}  Here, we use the same good parameters (provided by Ehsan Sadrfaridpour) for all instances. Tables~\ref{tab:comprunningtimes}--\ref{tab:time} as well as Figure~\ref{fig:distspeedups} summarize the results of our study. 

Looking at Table~\ref{tab:comp}, we first of all see that on 15 out of the 22 instances LPSVM computes a better or equal G-mean than mlsvm-AMG. On the remaining 7 instances our algorithm computes a result that is 0.04 worse on average.
On the \texttt{Isolet} instance, mlsvm-AMG computes a model that puts all data points on a single side. Hence, the G-mean value is zero on this instance. We exclude the instance from the following geometric mean computations.
Overall, the geometric average improvement in classification quality of LPSVM over mlsvm-AMG is $1\%$ when considering G-mean.  We therefore conclude that classification quality is comparable or often better than mlsvm-AMG.

Our fast configuration, LPSVM$_\text{fast}$ computes slightly worse models, but still has very good classification results. The geometric average in G-mean of our fast configuration is $3.2\%$ below the value of mlsvm-AMG.
Note that the fast and quality configuration of our algorithm use the same random seeds during coarsening and initial training. 
Still the classification quality on the input test data by LPSVM$_\text{fast}$ is sometimes better than the classification quality computed by the quality configuration, i.e., the configuration that uses additional refinement during uncoarsening. 
On first sight, this is somewhat surprising since the full algorithm returns the best model that been found during uncoarsening over all levels. 
However, this effect is due to the fact that our algorithm does not evaluate the model on the entire input data (since this would be too expensive), but uses a subset of the training set to perform the evaluations. Hence, the best model selected by the quality configuration is not necessarily the best model for the input training data.

\begin{table}[t]
  \centering
  \scriptsize
  \begin{tabular}{l||rrr||rr||r}
  &  \multicolumn{3}{c}{Speedup of}& \multicolumn{2}{c}{Speedup of} & \multicolumn{1}{c}{Speedup of}\\
Dataset & LPSVM &  LPSVM$_\text{fast}$ & LowDia  & LPSVM$_\text{par}$ & LowDia$_\text{par}$ & LPSVM$_\text{par}$\\
  & \multicolumn{3}{c||}{over mlsvm-AMG} &  \multicolumn{2}{c||}{over ThunderSVM} & \multicolumn{1}{c}{over LPSVM}\\

    \hline
\texttt{Advertisement}    & 2.37   & 11.18  & 1.31  & 1.27   & 1.51   & 0.79 \\
\texttt{APS failure}      & 16.84  & 61.54  & 6.86  & 103.36 & 58.28  & 2.80 \\
\texttt{Buzz}             & 0.42   & 6.78   & 0.30  & 94.24  & 143.30 & 0.64 \\
\texttt{Census}           & 12.67  & 79.97  & 4.90  & -      & -      & 1.48 \\
\texttt{Clean (Musk)}     & 1.78   & 3.63   & 0.73  & 6.44   & 3.05   & 1.70 \\
\texttt{Cod-rna}          & 3.34   & 21.50  & 7.08  & 38.32  & 116.24 & 0.56 \\
\texttt{EEG Eye State}    & 0.14   & 224.29 & 0.25  & 0.73   & 1.83   & 0.55 \\
\texttt{Forest (Class 3)} & 14.14  & 125.48 & 11.50 & -      & -      & 1.64 \\
\texttt{Forest (Class 5)} & 10.79  & 115.62 & 8.88  & 188.61 & 191.45 & 1.78 \\
\texttt{Forest (Class 7)} & 17.96  & 68.89  & 14.10 & -      & -      & 1.72 \\
\texttt{Hypothyroid}      & 0.81   & 3.26   & 1.54  & 0.41   & 1.18   & 0.81 \\
\texttt{Isolet (Class A)} & 64.58  & 218.14 & 44.33 & 23.09  & 17.59  & 2.68 \\
\texttt{Letter (Class A)} & 6.11   & 11.24  & 3.27  & 41.79  & 25.52  & 1.11 \\
\texttt{Letter (Class B)} & 15.31  & 34.20  & 8.09  & 29.11  & 18.88  & 1.00 \\
\texttt{Letter (Class H)} & 8.89   & 43.03  & 5.56  & 9.68   & 8.26   & 0.65 \\
\texttt{Letter (Class Z)} & 11.23  & 19.24  & 6.74  & 40.58  & 28.38  & 1.16 \\
\texttt{Nursery}          & 4.69   & 11.66  & 1.63  & 27.81  & 10.29  & 0.99 \\
\texttt{Protein}          & 122.64 & 263.03 & 24.62 & 113.65 & 43.49  & 1.06 \\
\texttt{Ringnorm}         & 0.67   & 20.87  & 2.38  & 6.49   & 3.51   & 3.83 \\
\texttt{Skin}             & 4.96   & 7.84   & 2.62  & 412.71 & 234.78 & 1.59 \\
\texttt{Sleep (Class 1)}  & 1.10   & 214.84 & 1.19  & 10.61  & 12.10  & 0.43 \\
\texttt{Twonorm}          & 12.19  & 17.98  & 3.33  & 21.86  & 6.14   & 0.88 \\
        \hline
        \hline
 geometric mean speedup   & 5.22   & 33.17  & 3.62  & 19.63  & 16.01  & 1.16 \\
        \hline
average G-score           & 0.90   & 0.87   & 0.91  & 0.90   & 0.91   & 0.90 \\

  \end{tabular}
  \caption{Speed-ups of our quality and fast algorithm configuration over
    mlsvm-AMG as well as our parallel algorithms over ThunderSVM on a per-instance basis. Instances with a '-' have not been solved by ThunderSVM within the time limit. In every case, average G-score is over all 22 instances. }
  \label{tab:time}
\end{table}

We now look at the running time spend to train the SVM by different algorithms. Figure~\ref{fig:distspeedups} shows the distribution of speed-ups of our algorithms over mlsvm-AMG, and Table~\ref{tab:comprunningtimes} and Table~\ref{tab:time} give detailed per-instance results. 
In terms of running time, already our higher-quality configuration LPSVM is faster on almost all instances. The exceptions are \texttt{Buzz}, \texttt{EEG Eye State}, \texttt{Hypothyroid} and \texttt{Ringnorm}.
In these cases, most of the time of our algorithm is spend in the refinement phase. The hierarchy of those instances yields a lot of support vectors for \emph{both} algorithms. That means that a lot of time is spend in solving the optimization problems on the different hierarchy levels. The better running time of the mlsvm-AMG solver is due to a partitioned training technique that we did not incorporate in our algorithm. This technique partitions the training set into $k$ blocks of roughly equal size in order to speedup computations. 
In general, \emph{positive} speed-ups for LPSVM range from $1.10$ up to two orders of magnitude. The largest observed speed-ups are on the instances \texttt{Isolet} (speed-up $64.58$) and \texttt{Protein} (speed-up $122.64$). 
The speedup on the \texttt{Protein} instance is due to the fact that our algorithm only has three levels in the multilevel hierarchy and there is only a very limited number of support vectors during uncoarsening. In contrast, the mlsvm-AMG algorithm has seven hierarchy levels on that instance. 
Sequentially, using low diameter clusterings improves classification quality, but slows down the training process. More precisely, the geometric mean speedup over mlsvm-AMG decrease from $5.22$ to $3.62$ when switching from label propagation clusterings to low diameter clusterings. This is due to the fact that this kind of clustering creates more levels in the hierarchy.
When considering the faster version of our algorithm LPSVM$_\text{fast}$, we observe that it is faster on every instance. This is not surprising since most of the time is spend during the refinement phase. However, since the observed classification quality of this approach is not much worse than mlsvm-AMG, we consider this a feasible alternative. For this configuration speed-ups range from $3.26$ up to $263.03$. The geometric average speed-up for the fast configuration over mlsvm-AMG is $33.17$.

Comparing to LibSVM, we note that our algorithms are orders of magnitude faster on all except once instance that LibSVM could solve within the time limit.
However, this is at the expense of solution quality, i.e., the solutions computed by LibSVM are mostly better than the ones computed by our algorithm. 
We now turn to the parallel algorithms and comparisons. Overall, it can be observed that ThunderSVM is mostly slower than mlsvm-AMG. This is not surprising since mlsvm-AMG already uses the multilevel idea to speed up LibSVM.
Note again that the main speedup of the multilevel technique here comes from the fact that model selection is done on the coarsest levels (and hence on much smaller instance) only. 
In contrast, tools like LibSVM and ThunderSVM perform model selection on the large input instance.
ThunderSVM is roughly an order of magnitude faster than LibSVM -- this is in line with the results reported by \cite{wenthundersvm18}.  Hence, when using ThunderSVM as a parallel solver during uncoarsening, our algorithm is speed up too. The maximum observed speedup is 3.83. However, since our algorithm is already fairly quick, the additional speedup on average is much smaller -- a factor 1.16. This is also partially the case, since ThunderSVM does on some instance slow down the overall process. However, LPSVM$_\text{par}$ still yields better classification quality than mlsvm-AMG.
In this case the largest speedup over mlsvm-AMG is a factor $173.09$ observed on instance \texttt{Isolet}. Overall, we conclude that using a parallel solver such as ThunderSVM can help to speed up computations, but this does not always have to be the case (see for example instances \texttt{EEG Eye State}, \texttt{Sleep (Class 1)}, or \texttt{Buzz}).
Already sequentially, our algorithm LPSVM is a factor $15.01$ faster than ThunderSVM (geometric mean speedup on the instances that ThunderSVM could solve within the time limit), while on average having the similar classification quality. Our parallel algorithm LPSVM$_\text{par}$ is overall a factor 19.63 faster than ThunderSVM.
In the parallel case, switching from label propagation clusterings to low diameter clusterings also has a positive impact on classification quality on average. 
Our fastest sequential algorithm, LPSVM$_\text{fast}$ is a factor of 88.65 faster than ThunderSVM, but classification quality deteriorates more.

\subsection{Experiments without Model Parameter Optimization}
\label{sec:largeinstances}

In this section, we perform experiments on the largest instances, i.e., \texttt{Forest}, \texttt{Mnist8M}, and \texttt{Susy}.
        Here, we compare different algorithms and configurations of our algorithm \emph{without} enabled model selection.
        Instead, we use predefined model parameters that have either been reported in the literature or came out of the previous section. In every case, all algorithm use the same model parameters.
        In order to gain additional speed up, we also use a sampling algorithm on the training data.
        That means that we compute a sample of the training data and train the model with the algorithm using the sample of $p\cdot|I|$ elements -- with $p \in (0,1)$ and $p$ is written next to the algorithm configuration.
        The final evaluation to estimate classification quality is done on~the~input~test~set.

The results are summarized in Table~\ref{tab:large_instances}. First we note that, even though no model selection is performed and fixed model parameters are used, ThunderSVM can only solve 3 instances within the 24 hour time limit.
On the instances that ThunderSVM could solve, our algorithms are faster by a large margin -- LPSVM$_\text{par}$ and LowDia$_\text{par}$ are on average (geometric mean) a factor 33.96 and 41.93 faster, respectively. Notably, classification quality is sometimes a lot better (see \texttt{Forest Class 5}) and sometimes slightly worse (see \texttt{Forest Class 7}).
As expected, using sampling typically worsens classification quality but also speedups the algorithm significantly.
For example, sampling 5\% of the input enables ThunderSVM to solve the largest three instances within the 24 hour time limit.
Still, our algorithms are still always faster with classification quality being on par -- for example on three instances LPSVM$_\text{par} 0.01$ computes better results while on the other three instances ThunderSVM$_\text{0.01}$ computes better results.
\begin{table*}[t]
  \scriptsize
  \centering
  \begin{tabular}{l||cc|rrrrrr|rrrrrr}
    {\multirow{2}{*}{Instances}} & {\multirow{2}{*}{C}} & {\multirow{2}{*}{$\gamma$}} & \multicolumn{2}{c}{LPSVM$_\text{par}$} & \multicolumn{2}{c}{LPSVM$_\text{par 0.01}$} & \multicolumn{2}{c|}{LPSVM$_\text{par 0.05}$}  \\
    & & & G-mean & s & G-mean & s & G-mean &  \\
    
    \hline
    \hline
\texttt{Forest (Class 3)}  & $2^{7.5}$  & $2^{-10}$   & 0.940 & 46.4     & 0.929 & 2.3   & 0.951 & 3.1    \\
\texttt{Forest (Class 5)}  & $2^{12.5}$ & $2^{-10.5}$ & 0.813 & 45.3     & 0.738 & 1.5   & 0.832 & 5.6    \\
\texttt{Forest (Class 7)}  & $2^{12.5}$ & $2^{0}$     & 0.959 & 1\ 397.9 & 0.578 & 12.7  & 0.821 & 49.4   \\
\texttt{Mnist8M (Class 0)} & 1000       & 0.006       & 0.960 & 2\ 445.5 & 0.960 & 207.0 & 0.956 & 248.5  \\
\texttt{Mnist8M (Class 1)} & 1000       & 0.006       & 0.948 & 2\ 382.6 & 0.949 & 209.5 & 0.957 & 209.5  \\
\texttt{Susy}              & $2^{15}$   & $2^{-7.5}$  & 0.752 & 579.2    & 0.723 & 18.4  & 0.713 & 39.4   \\
        \hline
        \hline
    {\multirow{2}{*}{Instances}} & {\multirow{2}{*}{C}} & {\multirow{2}{*}{$\gamma$}}   & \multicolumn{2}{c}{LowDia$_\text{par}$} & \multicolumn{2}{c}{LowDia$_\text{par 0.01}$} & \multicolumn{2}{c}{LowDia$_\text{par 0.05}$}  \\
    & & & G-mean & s & G-mean & s & G-mean & s \\
    
    \hline
    \hline
\texttt{Forest (Class 3)}  & $2^{7.5}$  & $2^{-10}$   &   0.951 & 46.9     & 0.939 & 2.2   & 0.948 & 3.3   \\
\texttt{Forest (Class 5)}  & $2^{12.5}$ & $2^{-10.5}$ &   0.876 & 55.6     & 0.757 & 2.3   & 0.847 & 5.9   \\
\texttt{Forest (Class 7)}  & $2^{12.5}$ & $2^{0}$     &   0.944 & 599.0    & 0.737 & 6.9   & 0.843 & 30.6  \\
\texttt{Mnist8M (Class 0)} & 1000       & 0.006       &   0.965 & 2\ 800.1 & 0.967 & 248.3 & 0.973 & 317.8 \\
\texttt{Mnist8M (Class 1)} & 1000       & 0.006       &   0.951 & 2\ 738.8 & 0.971 & 195.5 & 0.976 & 236.7 \\
\texttt{Susy}              & $2^{15}$   & $2^{-7.5}$  &   0.773 & 501.2    & 0.763 & 16.2  & 0.771 & 35.9  \\
\hline
\hline
    {\multirow{2}{*}{Instances}} & {\multirow{2}{*}{C}} &  {\multirow{2}{*}{$\gamma$}}&\multicolumn{2}{c}{ThunderSVM$_{full}$}  & \multicolumn{2}{c}{ThunderSVM$_{0.01}$}& \multicolumn{2}{c}{ThunderSVM$_{0.05}$}\\
    & & & G-mean & s & G-mean & s & G-mean & s \\
\hline
\hline
\texttt{Forest (Class 3)}   & $2^{7.5}$  & $2^{-10}$   & 0.889 & 4\ 964.0 &0.852 & 5.4   & 0.865 & 29.4      \\
\texttt{Forest (Class 5)}   & $2^{12.5}$ & $2^{-10.5}$ & 0.404 & 5\ 771.2 &0.343 & 4.5   & 0.290 & 28.8      \\
\texttt{Forest (Class 7)}   & $2^{12.5}$ & $2^{0}$     & 0.974 & 4\ 019.1 &0.522 & 49.1  & 0.803 & 104.7     \\
\texttt{Mnist8M (Class 0)}  & 1000       & 0.006       & -     & -        &0.993 & 264.1 & 0.998 & 1\ 750.0  \\
\texttt{Mnist8M (Class 1)}  & 1000       & 0.006       & -     & -        &0.994 & 427.5 & 0.998 & 1\ 784.0  \\
\texttt{Susy}               & $2^{15}$   & $2^{-7.5}$  & -     & -        &0.779 & 542.4 & 0.781 & 3\ 913.4  \\

  \end{tabular}
 \caption{Results of different algorithms on large instances using fixed model parameters.}\label{tab:large_instances}
\end{table*}

\section{Conclusion}
We present a very fast (parallel) multilevel support vector machine that uses a label propagation or a low diameter clustering algorithm to
 construct a problem hierarchy.
 In contrast to previous methods such as LibSVM or ThunderSVM, our algorithm uses a multilevel scheme to enable quick model selection and fast training.
In contrast to previous multilevel approaches, our clustering contraction scheme is able to shrink the input size very quickly and hence to build the hierarchy much quicker than before.
Extensive experiments indicate that our algorithm is less affected by the number of features of the data set.
Moreover, our new algorithm achieves speed-ups up multiple orders of magnitude while having similar or better classification quality over state-of-the-art algorithms.
Our implementation is publicly available in the open source framework~KaSVM\footnote{\url{https://algo2.iti.kit.edu/kasvm/}}.
Important future work includes to incorporate automatic algorithm configuration tools such as~\citep{DBLP:journals/jair/HutterHLS09} that are able to predict good algorithm parameters based on properties of the input instance.

\renewcommand{\bibname}{\begin{flushleft} References \end{flushleft}}
\bibliography{phdthesiscs,literatur}
\end{document}